\documentclass[letterpaper, 10pt, conference]{ieeeconf}      % Use this line for a4
\IEEEoverridecommandlockouts                              % This command is only
\usepackage{times}

\usepackage{multicol}
\usepackage[bookmarks=true]{hyperref}

\usepackage{gensymb}
\usepackage{tabularx}
\usepackage{graphicx}
\usepackage{caption}
\usepackage{subfig}

\usepackage{amsmath}
\usepackage{amssymb}
\usepackage{svg}
\usepackage[ruled,vlined]{algorithm2e}
\usepackage{algorithmic}
\usepackage{graphicx}
\usepackage{cite}
\usepackage{float}
\usepackage{xcolor}

\newcommand{\argmin}{\operatornamewithlimits{argmin}}

\usepackage{xargs}                      % Use more than one optional parameter in a new commands
\begin{document}

\title{\LARGE \bf
RadarSLAM: Radar based Large-Scale SLAM in All Weathers}

% You will get a Paper-ID when submitting a pdf file to the conference system
\author{Ziyang Hong, Yvan Petillot and Sen Wang% <-this % stops a space
\thanks{The authors are with Edinburgh Centre for Robotics, Heriot-Watt University, Edinburgh, EH14 4AS, UK.
        {\tt\small \{zh9, y.r.petillot, s.wang\}@hw.ac.uk}}%
}

%\author{\authorblockN{Michael Shell}
%\authorblockA{School of Electrical and\\Computer Engineering\\
%Georgia Institute of Technology\\
%Atlanta, Georgia 30332--0250\\
%Email: mshell@ece.gatech.edu}
%\and
%\authorblockA{Twentieth Century Fox\\
%\and
%\authorblockN{James Kirk\\ and Montgomery Scott}
%\authorblockA{Starfleet Academy\\
%San Francisco, California 96678-2391\\

% avoiding spaces at the end of the author lines is not a problem with
% conference papers because we don't use \thanks or \IEEEmembership

%Homer Simpson\authorrefmark{2},
%Montgomery Scott\authorrefmark{3} and
%Eldon Tyrell\authorrefmark{4}}
%Telephone: (800) 555--1212, Fax: (888) 555--1212}
%\authorblockA{\authorrefmark{4}Tyrell Inc., 123 Replicant Street, Los Angeles, California 90210--4321}}

\maketitle

\begin{abstract}
Numerous Simultaneous Localization and Mapping (SLAM) algorithms have been presented in last decade using different sensor modalities. However, robust SLAM in extreme weather conditions is still an open research problem. In this paper, RadarSLAM, a full radar based graph SLAM system, is proposed for reliable localization and mapping in large-scale environments. It is composed of pose tracking, local mapping, loop closure detection and pose graph optimization, enhanced by novel feature matching and probabilistic point cloud generation on radar images. Extensive experiments are conducted on a public radar dataset and several self-collected radar sequences, demonstrating the state-of-the-art reliability and localization accuracy in various adverse weather conditions, such as dark night, dense fog and heavy snowfall.

% verify the performance of our system in two setup: radar odometry only and full SLAM system on public avaliable dataset. We also collect dataset in extreme weathers, our experiments show that our system is robust in various weather conditions, such as dark night, dense fog and heavy snowfall. We demonstrate the versatility and reliability of using radar for tackling the long-term navigation problem for autonomous vehicles.
\end{abstract}

\IEEEpeerreviewmaketitle

\section{Introduction}
Simultaneous Localization and Mapping (SLAM) has been extensively investigated with numerous sensor modalities, e.g., sonar, camera and LiDAR, in the last decades. However, for outdoor large-scale SLAM, ensuring its robust operation is still very challenging especially in adverse weather conditions. Recently, the emerging Frequency-Modulated Continuous Wave (FMCW) radar sensors which can work in various weathers have been increasingly adopted for self-diving cars and autonomous robots. Therefore, an interesting yet open question is whether these radars can be used for robust SLAM in large-scale environments in extreme weather conditions, such as heavy snowfall.

Some radar landmark extraction and motion estimation approaches are proposed in \cite{marck2013indoor,vivet2013localization,schuster2016landmarkRadar}. In \cite{holder2019realRadarGraph}, radar scans are represented as point clouds and relative motion is estimated by an Iterative Closest Point (ICP) algorithm. Feature based geometric methods are also investigated for radar odometery \cite{cen2018precise,cen2019radar,aldera2019Introspective}. For instance, a dedicated radar landmark extraction algorithm is proposed in \cite{cen2018precise} by estimating the radar power signal. In \cite{cen2019radar}, a graph matching algorithm is designed for radar keypoint data association without the need of motion prior. Recently, a direct radar odometry method is proposed to estimate relative pose using Fourier Mellin Transform and local graph optimization \cite{PhaRaO}. 

\begin{figure}[t!]
    \centering
    \includegraphics[width=0.95\columnwidth]{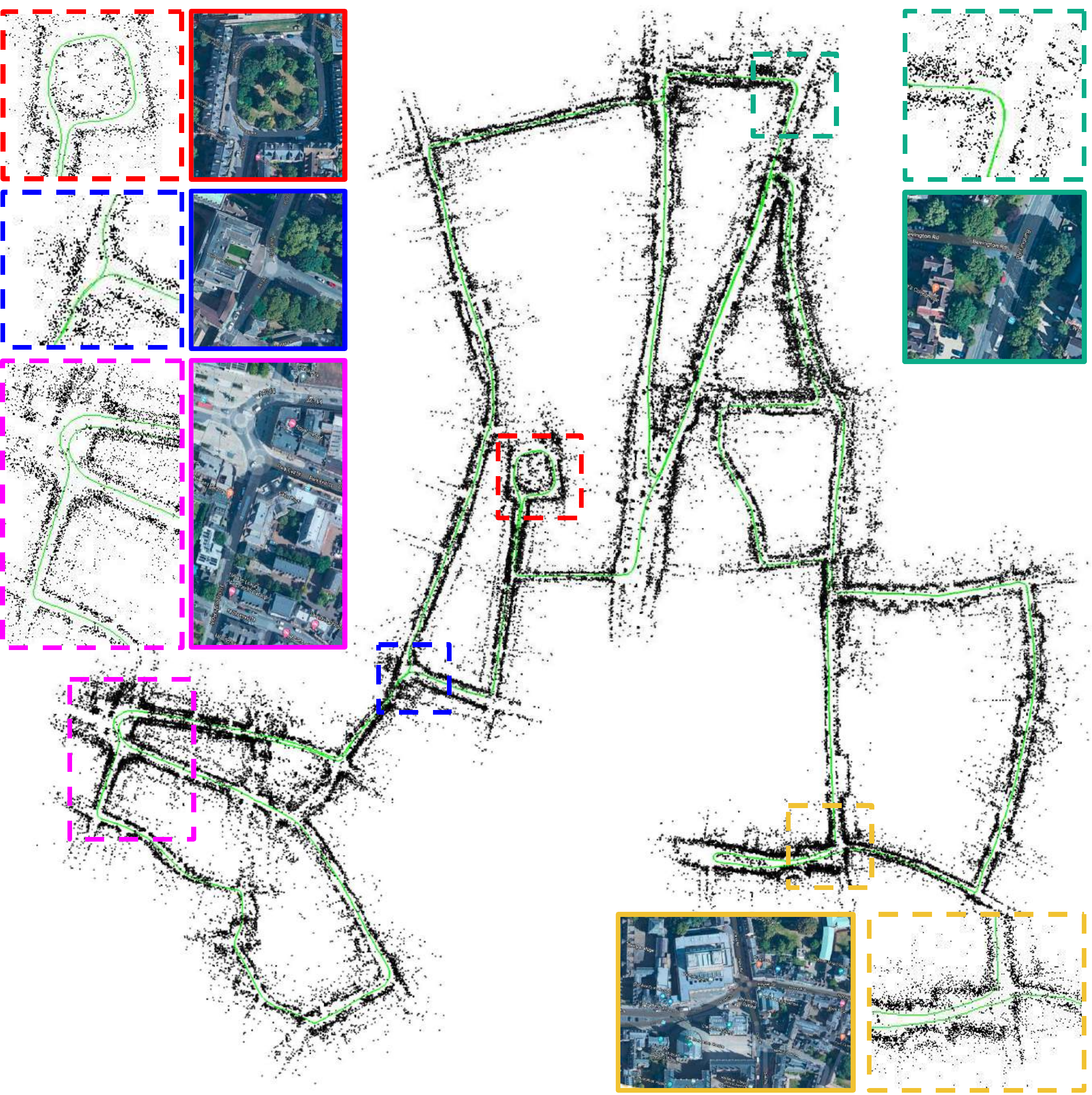}
    \caption{Mapping Result of the proposed RadarSLAM on Oxford Radar Dataset (Sequence 10-12-32-52) \cite{RadarRobotCarDatasetICRA2020}. Green line shows the estimated trajectory. Total length is 9.04 km.}
    % Bottom right: mapping result on sequence \textbf{Snow} of self-collected dataset, this sequence is \textbf{8.7 Km}. Bottom middle: radar image. Bottom left: camera images.}
    % \sen{another example in snow and snow images and radars}
    \label{fig:map}
\end{figure}

Deep Learning based radar feature extraction and odometry approaches have also been explored \cite{barnes2019masking,aldera2019fast,UnderTheRadarICRA2020}. Specifically, in \cite{aldera2019fast} the coherence of multiple measurements is learnt to decide which information to be kept in the reading while a mask is generated to filter out the noises from the radar reading \cite{barnes2019masking}. A self-supervised framework is proposed for robust keypoint detection on Cartesian radar images which is further used for both motion estimation and loop closure detection \cite{UnderTheRadarICRA2020}.

% Two datasets \cite{RadarRobotCarDatasetICRA2020} \cite{mulran} using FMCW radar have been recently released. MulRan \cite{mulran} datasat provides both dense Lidar and long range radar data covering various environments in different sequences. MulRan focus on providing structural diversity and temporal diversity. It has numerous revisits within sequence which is suitable for evaluation of place recognition task. Oxford Radar RobotCar dataset \cite{RadarRobotCarDatasetICRA2020} traverses the same route within the city of Oxford for 32 times with longer distance and larger scale with additional sensor like stereo cameras and side cameras. \cite{mulran} provides 6 Degree-of-Freedom ground truth trajectory while \cite{RadarRobotCarDatasetICRA2020} gives 3 Degree-of-Freedom ground truth poses.

% we explore how to achieve radar SLAM in large-scale outdoor environments.
In this paper, a RadarSLAM system is proposed for robust localization and mapping in large-scale outdoor environments by making use of radar geometry and graph SLAM. Our main contributions include:
\begin{itemize}
    \item An efficient yet reliable feature matching and pose tracking algorithm using radar geometry and graph representation.
    \item A probabilistic point cloud generation from radar images dramatically reducing speckle noises.
    \item Graph optimization based full radar SLAM system which can operate even in adverse weather conditions.
    \item Extensive real experiments in large-scale environments, demonstrating the first time reliable radar SLAM in extreme weathers, e.g., dense fog and heavy snowfall.
\end{itemize}

The rest of this paper is organized as follows. Section \ref{section:radar_background} introduces background on radar sensing, followed by algorithm description on RadarSLAM in Section \ref{sec:radar_slam}. Experimental results are given in Section \ref{sec:exp_results} before drawing the conclusions.

\section{Radar Sensor Background}
\label{section:radar_background}

\begin{figure}[t]
    \centering
    \includegraphics[width=0.7\columnwidth]{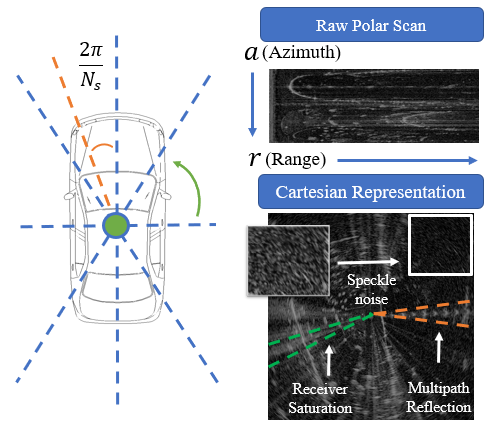}
    \caption{Polar and Cartesian Representations of a 360\degree FMCW. Three common noises of a radar image are speckle noise, receiver saturation and multi-path reflection.}
    \label{fig:fmcw_radar}
\end{figure}

FMCW radar is a type of radar whose transmitter sends waveforms and receiver waits for the echo reflected from the targets. It can not only measure the distance between the radar and the target, but also predict the target's velocity by analyzing the frequency difference between the transmitted and received signals \cite{millimeterFMCW}. However, the current FMCW radar sensor suffers from multiple sources of noises, e.g., range error, angular error and false-positive and false-negative detections \cite{cen2019ego}. For example, the false-positive detections include clutter, sidelobes, multi-path reflections and receiver saturation since the sensor is highly sensitive to the surface reflectance and the reflector pose. Multi-path reflection can cause inconsistency between consecutive frames, introducing extra noises and outliers. Therefore, radar readings tend to be noisier than the camera and LiDAR data, making them harder to work with for motion estimation and SLAM.

\textbf{Radar Geometry} A 360\degree FMCW radar sweeps continuously with a total of $N_s$ azimuth angles for a full 360\degree as shown in Fig. \ref{fig:fmcw_radar}, i.e., the step size on azimuth angle is $2\pi/N_s$. For each azimuth angle, the radar emits a beam and collapses the return signal as a range distance without considering elevation. Therefore, a radar image can provide absolute metric information of range distance, different from a camera image which loses depth. The raw polar scan can be transformed into a Cartesian space, being represented by a grey scale image. Therefore, given a point $(a,r)$ on the polar image, where $a$ and $r$ denote the azimuth and range respectively, its Cartesian coordinate $\mathbf{P}$ can be computed by
\begin{equation}
    \mathbf{P} = \begin{bmatrix} \gamma\cdot r\cdot\cos\theta
    \\ \gamma \cdot r\cdot \sin \theta \end{bmatrix}
\end{equation}
where $\theta = 2\pi\cdot a/ N_s$ is the ranging angle in the Cartesian coordinate, and $\gamma$ is the scaling factor between the image pixel space and the world metric space. For better resolution, the Cartesian image is usually interpolated using bi-linear interpolation. This work employs the Cartesian representation for easier understanding.

\section{Radar based SLAM}\label{sec:radar_slam}

% \subsection{Notation}
% \subsection{Large-Scale Radar SLAM}

\begin{figure}
    \centering
    \includegraphics[width=\columnwidth]{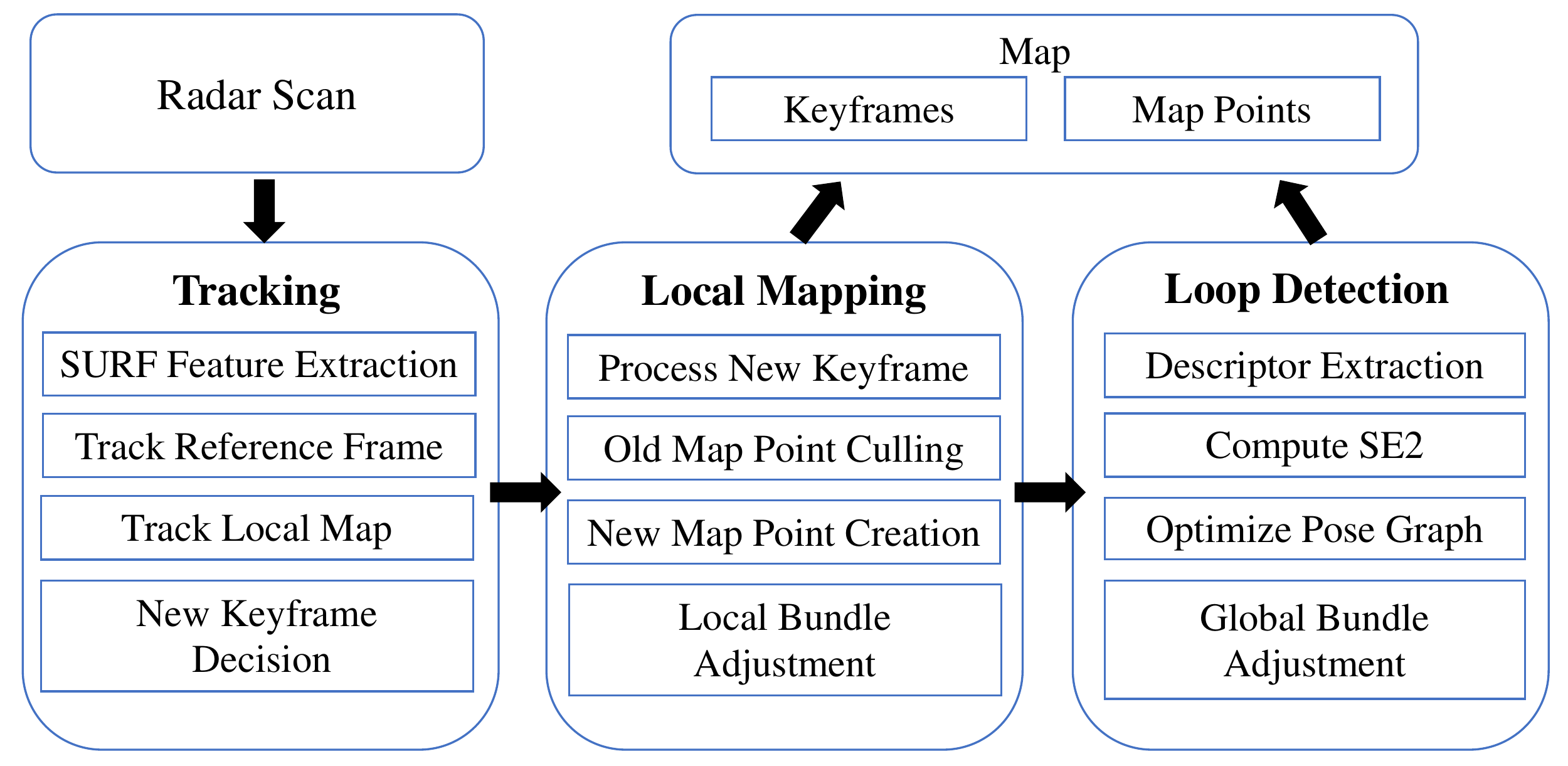}
\caption{System Diagram}
\label{fig:system_diagram}
\end{figure}

Given a sequence of radar scans, RadarSLAM aims to estimate radar (robot) poses and a global consistent map using graph SLAM. To this end, the proposed RadarSLAM system is designed to have four main subsystem: pose tracking, local mapping, loop closure detection and pose graph optimization. The system overview is shown in Fig. \ref{fig:system_diagram}.

\begin{figure*}[h]
    \includegraphics[width=\textwidth]{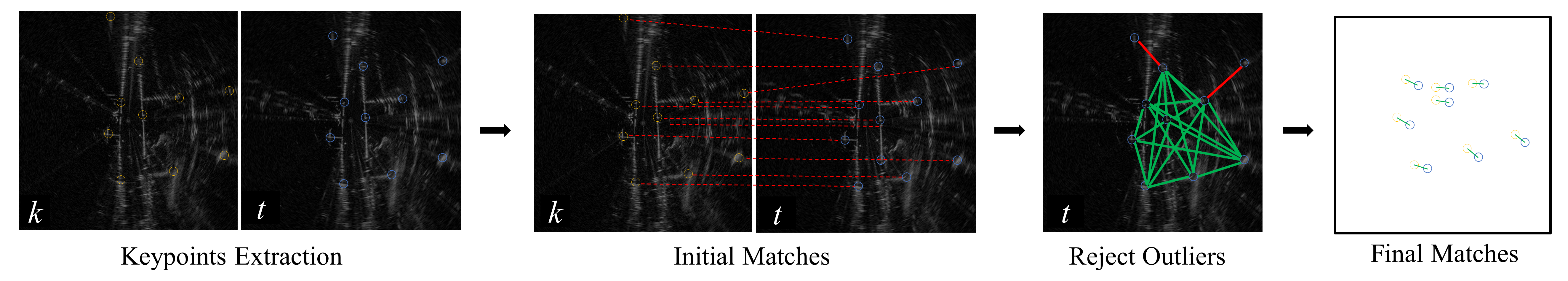}
    \caption{Feature Extraction and Matching using Outlier Rejection. The yellow and blue circles are keypoints detected. The initial matches are within a certain radius of the query keypoints. The green graph is the maximum clique where every node is inter-connected. The red connections indicate the false matches that are not part of the maximum clique and rejected.}
    \label{fig:keypoint_matching}
\end{figure*}

\subsection{Pose Tracking}

Pose tracking continuously estimates the radar pose online in the framework of keyframe based pose tracking. Specifically, to track the pose $\mathbf{C}_t$ of the current radar frame $t$ in the world coordinate system, the relative transformation $\mathbf{T}_{tk} \in \textbf{SE}(2)$ between the current frame $t$ and the keyframe $k$ with pose $\mathbf{C}_k$ needs to be computed. $\textbf{SE}(2)$ represents special Euclidean group.
% $R_{ij} \in  \textbf{SO}(2)$ and $t_{ij} \in \mathbb{R}^2$ is the rotation part and the translation part of $T_{ij}$ respectively.
Then, assume the keyframe pose $\mathbf{C}_k$ is known, $\mathbf{C}_t$ can be computed by
\begin{equation}
  \mathbf{C}_t = \mathbf{C}_k \mathbf{T}_{tk}^{-1} \label{eq:transformation}
\end{equation}

The geometry of keypoints of the Cartesian radar images is exploited to calculate $\mathbf{T}_{tk}$. Two sets of keypoint features are extracted from the current frame $t$ and keyframe $k$ respectively using a feature extraction algorithm, e.g., SURF \cite{bay2008speeded}. Then, these two sets of keypoints are matched using their feature descriptors. Different from vision based methods, feature matching of radar images can be elaborated by making full use of radar geometry which directly provide metric ranging information. Therefore, two mechanisms are adopted to reduce the number of incorrect feature matches for radar images.
% The Cartesian radar image  coordinate, keypoints are extracted from the Cartesian radar image and descripted by using . The keypoint feature sets of the keyframe and current frame are denoted as $F_j$ and $F_i$ respectively.
First, motion prior (e.g., maximum velocity) is introduced to restrict the maximum searching radius of correspondences for a query keypoint on the radar local coordinate system. This decreases both the number of incorrect matches and computation time needed for exhaustive feature matching.
% Denote the initial feature matches as $M_{ij}$ between frame $i$ and keyframe $j$ and a pair of match is denoted as ($f_a^i$, $f_a^j$).
Second is a pairwise consistency constraint, which further rejects outliers considering the fact that the pairs of inliner keypoint correspondences should follow a similar motion tendency. Therefore, for any two pairs of keypoint matches between the current frame $t$ and the keyframe $k$, they should satisfy the following pairwise constraint:
\begin{equation}
  \left|\lVert\mathbf{P}_t^i - \mathbf{P}_k^i\rVert_2 - \lVert\mathbf{P}_t^j - \mathbf{P}_k^j\lVert_2\right| < \delta_c
\end{equation}
where $\left|\cdot\right|$ is the absolute operation, $\lVert\cdot\lVert_2$ is the Euclidean distance, $\mathbf{P}_t^i$ and $\mathbf{P}_k^i$ are the Cartesian coordinates of the keypoint pair $i$ in the local coordinate system, $\mathbf{P}_t^j$ and $\mathbf{P}_k^j$ are these of the keypoint pair $j$, and $\delta_c$ is a small distance threshold. A consistency matrix $\mathbf{G}$ is then used to represent all the matches that satisfy this pairwise consistency. If a pair of matches satisfies this constraint, the corresponding entry in $\mathbf{G}$ is set as 1. Finding the maximum inlier set that the matches are mutually consistent is equivalent of deriving the maximum clique of a graph represented by $\mathbf{G}$. Once the maximum inlier set is obtained, its keypoint matches are used to compute the relative transformation $\mathbf{T}_{tk}$ by using Singular Value Decomposition (SVD) \cite{challis1995procedure}. Following Eq. \ref{eq:transformation}, the current pose $\mathbf{C}_{t}$ can be obtained as an initial guess for the optimization in Eq. \ref{eq:projection_error}. An example of the feature exaction and matching is given in Fig. \ref{fig:keypoint_matching}.

In order to further constrain local drifts, the final $\mathbf{C}_{t}$ is obtained by minimizing the reprojection errors of the keypoint pairs successfully matched between the current frame $t$ and the keyframe $k$:
\begin{equation}\label{eq:projection_error}
  \mathbf{C}_t^* = \argmin_{\mathbf{C}_t} \sum_{i \in M_{tk}} \lVert\mathbf{P}_w^i - \mathbf{C}_t\mathbf{P}_t^i\rVert_2
  % _{f^i_a,f^j_a  \in M_{ij}}_{f^i_a,f^j_a  \in M_{ij}}
\end{equation}
where $i$ is the $i$th feature matching pair in all the pairs $M_{tk}$ between the current frame $t$ and the keyframe $k$, and $\mathbf{P}_w^i$ is the coordinate of its corresponding map point in the world coordinate system.

As the current frame is processed, it may be converted as a keyframe for better tracking robustness and accuracy. We follow the similar criteria used in visual SLAM \cite{murTRO2015} for keyframe generation, i.e., considering the conditions on the minimum number of keypoint matches, the translation and rotation between the current frame and and the keyframe.

\subsection{Local Mapping}

The goal of local mapping is to refine the pose estimation and local map consistency by jointly optimizing the poses and local map estimated. It runs on parallel with the pose tracking thread. Once a new keyframe is created, its keypoints are created as map points in the world coordinate system. Then, the nearby keyframes and the map points which can be observed by these keyframes are retrieved to perform local Bundle Adjustment \cite{bundle}, i.e., the poses of the keyframes and the locations of the map points are optimized by minimizing the weighted Sum of Squared Error cost function:
\begin{equation}
    \mathbf{X}^* = \argmin_{\mathbf{X}} \frac{1}{2}\sum_{i}(\mathbf{\hat{z}}_i - \mathbf{z}_i(\mathbf{X}))^\top \mathbf{W}_i (\mathbf{\hat{z}}_i - \mathbf{z}_i(\mathbf{X}))
\end{equation}
where $\mathbf{X}$ is the state of both keyframe poses and map point locations, $\mathbf{\hat{z}}_i - \mathbf{z}_i(\mathbf{X})$ is the residual error between the prediction and observation of the $i$th map point, and $\mathbf{W}_i$ is a symmetric positive definite weighting matrix. This optimization is solved by using the Levenberg-Marquadt method. In order to limit the computation needed, the map points that created by last keyframe are culled if they are not observable by more than two keyframes.

% The poses of the keyframes and the locations of the map points are jointly optimized

% \begin{figure}
%     \centering
%     \begin{subfigure}[b]{0.32\columnwidth}
%     \includegraphics[width=\columnwidth]{figures/system/loop_origin_image.jpg}
%     \caption{}
%     \end{subfigure}
%     \begin{subfigure}[b]{0.32\columnwidth}
%     \includegraphics[width=\columnwidth]{figures/system/loop_naive_pointcloud.jpg}
%     \caption{}
%     \end{subfigure}
%     \begin{subfigure}[b]{0.32\columnwidth}
%     \includegraphics[width=\columnwidth]{figures/system/loop_filtered_pointcloud.jpg}
%     \caption{}
%     \end{subfigure}
%     \caption{Peaks Detection of a Radar Scan. (a): Original Cartesian image. (b): Peaks (in yellow) detected using a local maxima algorithm. Note a great amount of peaks are speckle noises. (c): Peaks detected using the proposed point cloud extraction algorithm which preserves the environmental structure and suppresses the detections from the multi-path reflection and speckle noises.}
%     \label{fig:loop_closure}
% \end{figure}
\begin{figure}
    % \centering
    \subfloat{}
    {\includegraphics[width=0.32\columnwidth]{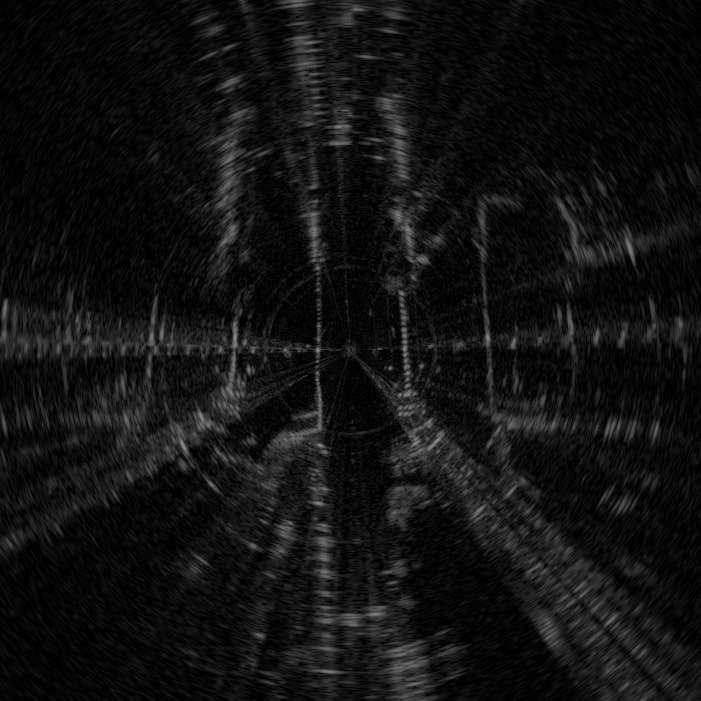}}
    \hfill
    \subfloat{}
    {\includegraphics[width=0.32\columnwidth]{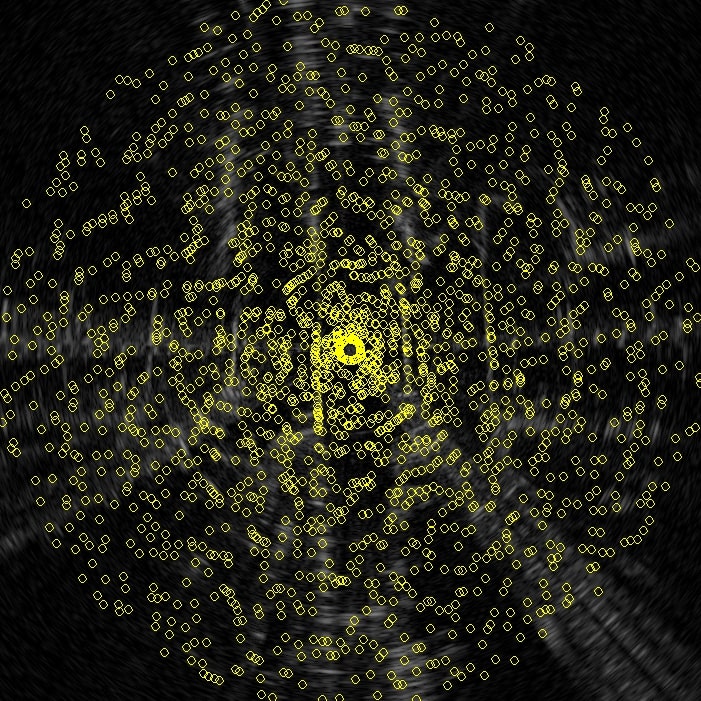}}
    \hfill
    \subfloat{}
    {\includegraphics[width=0.32\columnwidth]{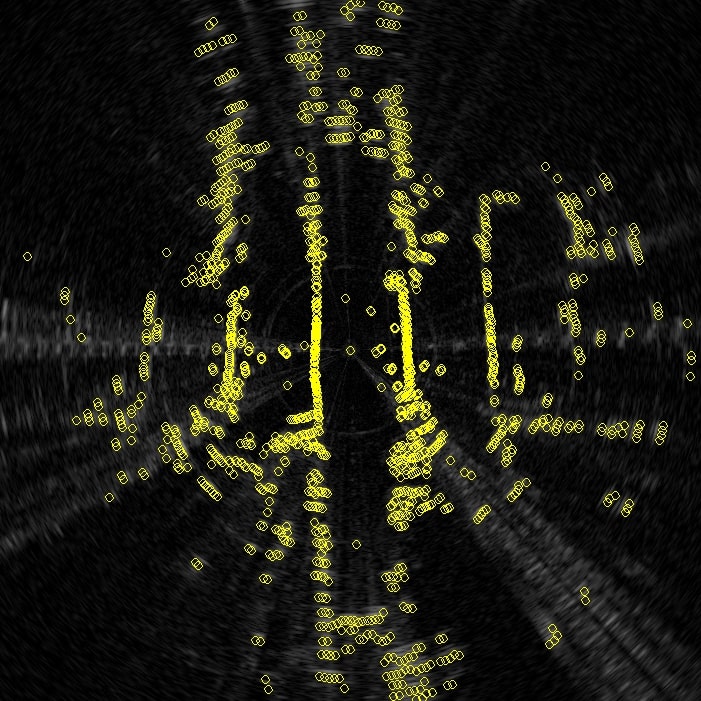}}

    \caption{Peaks Detection of a Radar Scan. (a): Original Cartesian image. (b): Peaks (in yellow) detected using a local maxima algorithm. Note a great amount of peaks are speckle noises. (c): Peaks detected using the proposed point cloud extraction algorithm which preserves the environmental structure and suppresses the detections from the multi-path reflection and speckle noises.}
    \label{fig:loop_closure}
\end{figure}

\subsection{Loop Closure Detection}

\begin{table*}[t!]
  \centering
  \caption{Results on Oxford Radar RobotCar Dataset}
  \label{tab:result_oxford}
  \begin{tabular}{ l|cccccc  }
  \hline
  & \multicolumn{6}{c}{\textbf{Sequence}} \\
  \textbf{Method} & 10-12-32-52 & 16-13-09-37 & 17-13-26-39 & 18-14-14-42 & 18-15-20-12 & Mean \\
  \hline
  Visual Odometry Baseline \cite{churchill2012experience} &N/A &N/A &N/A &N/A &N/A & 3.9802/0.0102 \\
  ORB-SLAM2 Stereo \cite{murTRO2015} &66.94/0.2019 &68.81/0.2186 &66.68/0.2175 &63.92/0.1978 &65.06/0.1877 &66.3081/0.2047   \\
  \hline
  LOAM \cite{loamZhang}  &75.56/0.2051 &58.97/0.2158 &54.10/0.1804 &56.24/0.1964 &67.30/0.2290 &62.43/0.2053 \\
  SuMa \cite{behley2018rss}  &78.32/0.1811 & 75.96/0.1617 &89.05/0.028 &77.98/0.1792 &77.81/0.1712 & 79.82/0.1081 \\
  \hline
  % Cen Full Resolution\cite{cen2018precise}  &N/A &N/A &N/A &N/A &N/A &8.4730/0.0236\\
  Cen 0.1752 m/pixel \cite{cen2018precise}  &N/A &N/A &N/A &N/A &N/A &3.7168/0.0095\\
  % Barnes Dual Polar\cite{barnes2019masking}  &N/A &N/A &N/A &N/A &N/A &1.2621/0.0036\\
  Barnes Dual Cart \cite{barnes2019masking} &N/A &N/A &N/A &N/A &N/A &1.1627/0.0030\\
  Our Radar Odometry & 2.9899/0.0086 &3.1202/0.0093 &2.9293/0.0087 &3.1886/0.0095 &2.8585/0.0092 &3.0173/0.0091 \\
  Our Radar SLAM     & 2.1661/0.0067 &1.8382/0.0059 &2.4596/0.0081 &2.2152/0.0071 &2.2478/0.0078 & 2.1854/0.0071 \\
  \hline
  \end{tabular}\par
% \bigskip
Results are given as \textit{translation error} / \textit{rotation error}. Translation error is in \%, and rotation error is in degrees per meter (deg/m).
\end{table*}

Robust loop closure detection is critical to reduce drifts for a SLAM system. Although Bag-of-Words model has been proved efficient for vision based SLAM algorithms, it is not adequate for radar based loop closure detection due to three main reasons: First, radar images have less distinctive characteristics on pixels compared with the optical images, which means similar feature descriptors can be repeated widely across radar images; Second, the multi-path reflection problem in radar can introduce ambiguity for the feature descriptor; Third, a small rotation of the radar sensor may produce tremendous scene changes, significantly distorting the histogram distribution of the descriptors. Therefore, we adopt a technique which captures the scene structure and exploit the spatial signature of the reflection density from radar point clouds.
% Large part of the structure in the scene will disappear and new structure will be visible at the same time

\begin{algorithm}[!t]
  \SetAlgoLined
  \textbf{Input:} Radar image $S \in \mathbb{R}^{m \times n} $\;
  \textbf{Output:} Point Cloud $P \in \mathbb{R}^{z \times 2}$\;
  \textbf{Parameters:} Minimum peak prominence $\delta_{p}$ and minimum peak distance $\delta_{d}$\;
  Initialize empty point cloud set $P$\;
      \For{$i \gets 1$ to $m$}{
      $Q^{kx1}$ $\leftarrow$ findPeaks($S[i,:]$, $\delta_{p}$,  $\delta_{d}$)\;
      $(\mu, \sigma)$ $\leftarrow$ meanAndStandardDeviation($Q^{k \times 1}$)\;
          \For{each peak $q$ in $Q$}{
              \If{$q \geq (\mu + \sigma)$}
              {
                  $p$ $\leftarrow$ transformPeakToPoint($q$, $i$)\;
                  Add the point $p$ to $P$\;
              }
          }
      }
  % \For{$i \gets 1$ to $m$}
  %     K \leftarrow findPeaks(a, \delta_{p},  \delta_{d})
  % \EndFor
      \caption{Radar Image to Point Cloud Conversion}
      \label{algo:point_cloud_generation}
\end{algorithm}

The radar images are first converted as point clouds. An intuitive and naive way would be detecting peaks by finding the local maxima from each azimuth reading. However, as shows in Fig. \ref{fig:loop_closure}, the peaks can be distributed randomly across the whole radar image, even for the areas with no real object, due to the speckle noises. Therefore, we propose a simple yet effective point cloud generation algorithm using probabilistic model. Assume that the peak power $s$ in each azimuth scan follows a normal distribution as
\begin{equation}
    f(s) = \frac{1}{\sqrt{2\pi}\sigma} \exp\left(-\frac{(s-\mu)^2}{2\sigma^2} \right)
\end{equation}
where $\mu$ and $\sigma$ are the mean and the standard deviation of the powers of the peaks in one azimuth scan. By selecting the peaks lie beyond one standard deviation and are greater than their mean, the true detection can be separated from the false-positive ones. The procedure is shown in Algorithm \ref{algo:point_cloud_generation}. Once a point cloud is generated from a radar image, M2DP \cite{he2016m2dp}, a rotation invariant global descriptor designed for 3D point clouds, is adopted to describe it for loop closure detection. M2DP computes the density signature of the point cloud on the plane and uses the left and right singular vectors of these signatures as the descriptor.

\begin{figure*}
    \centering
    \subfloat[10-12-32-52]{\includegraphics[width=0.19\textwidth]{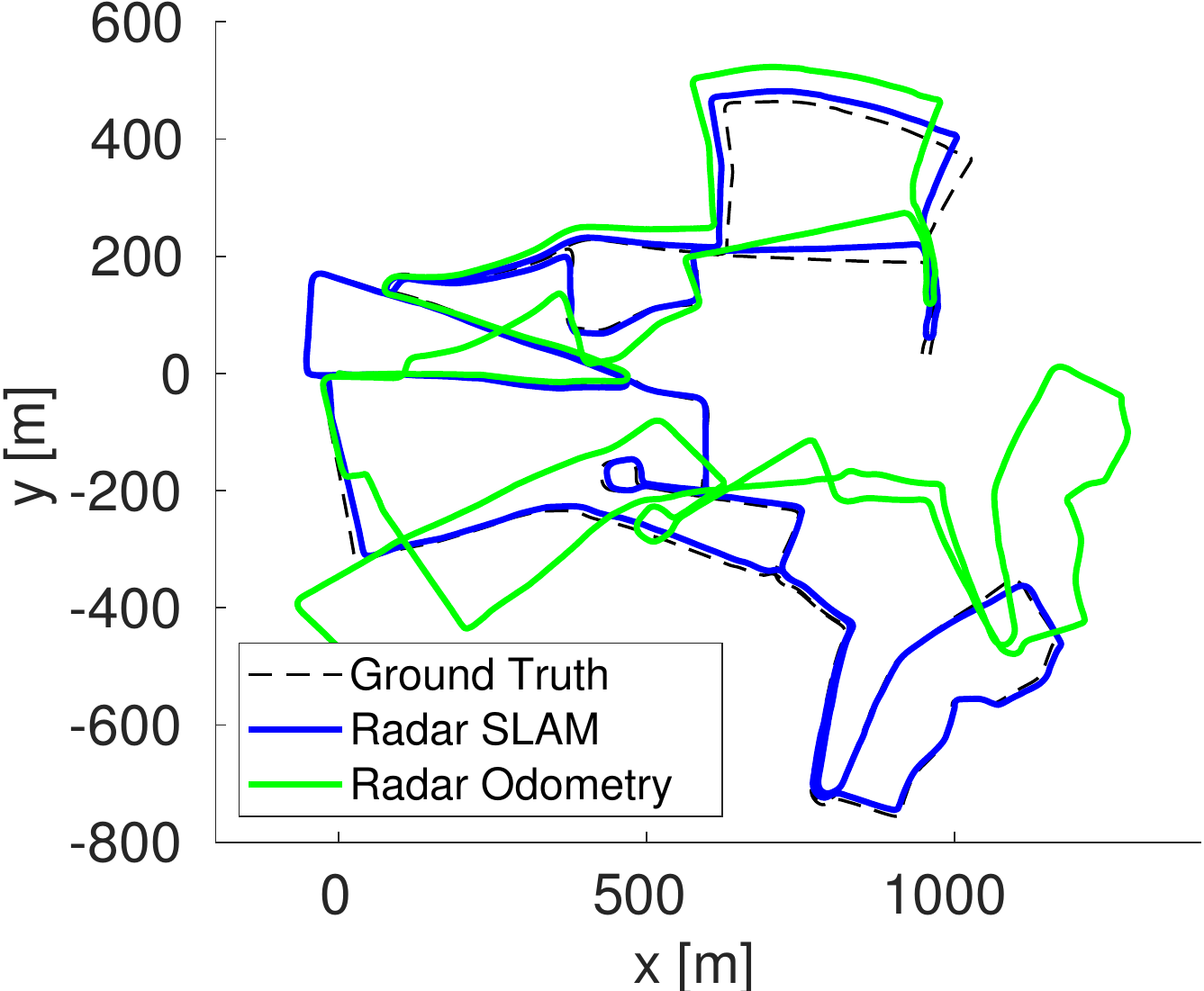}}\,
    % \hfill
    \subfloat[16-13-09-37]{\includegraphics[width=0.19\textwidth]{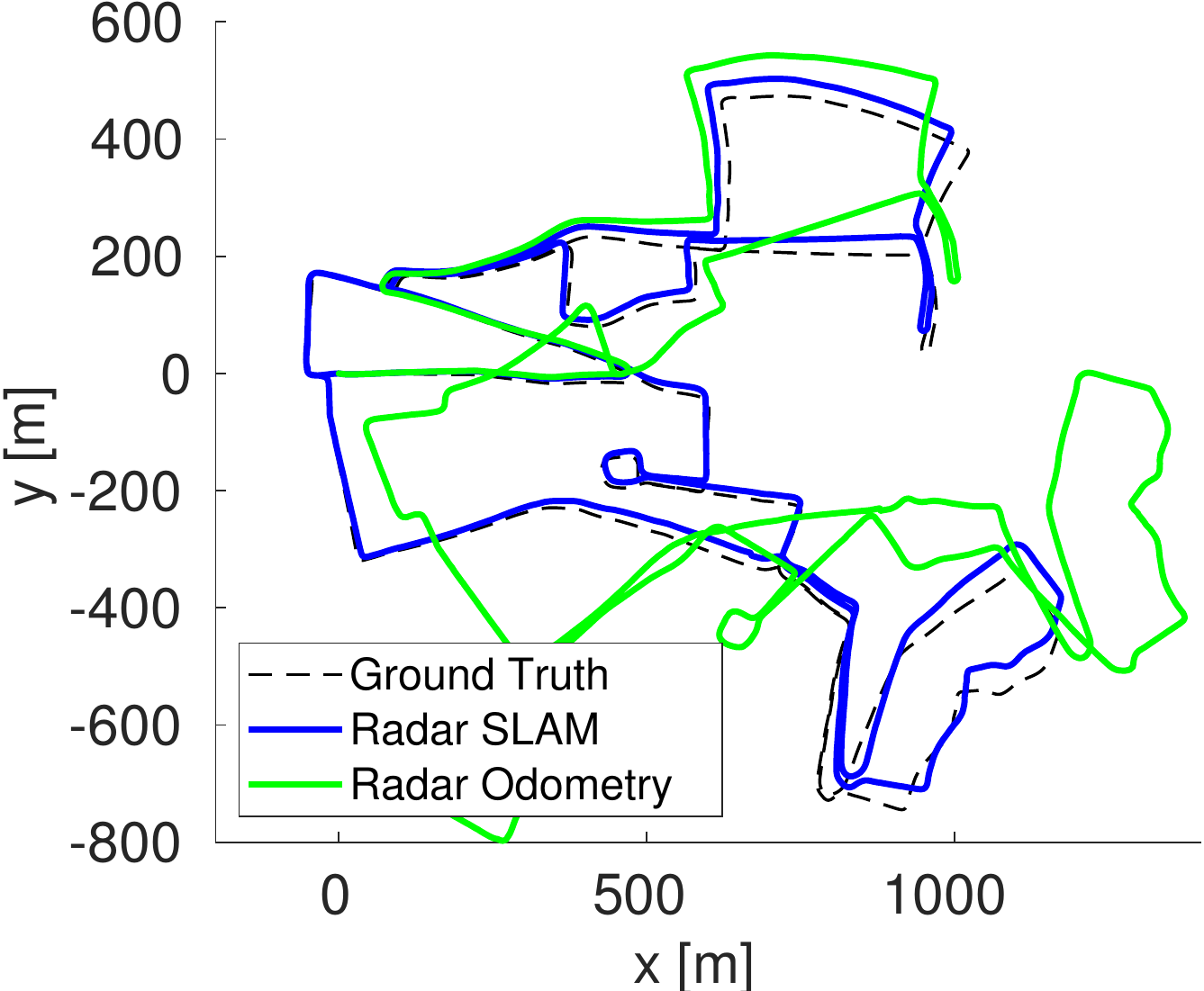}}\,
    % \hfill
    \subfloat[17-13-26-39]{\includegraphics[width=0.19\textwidth]{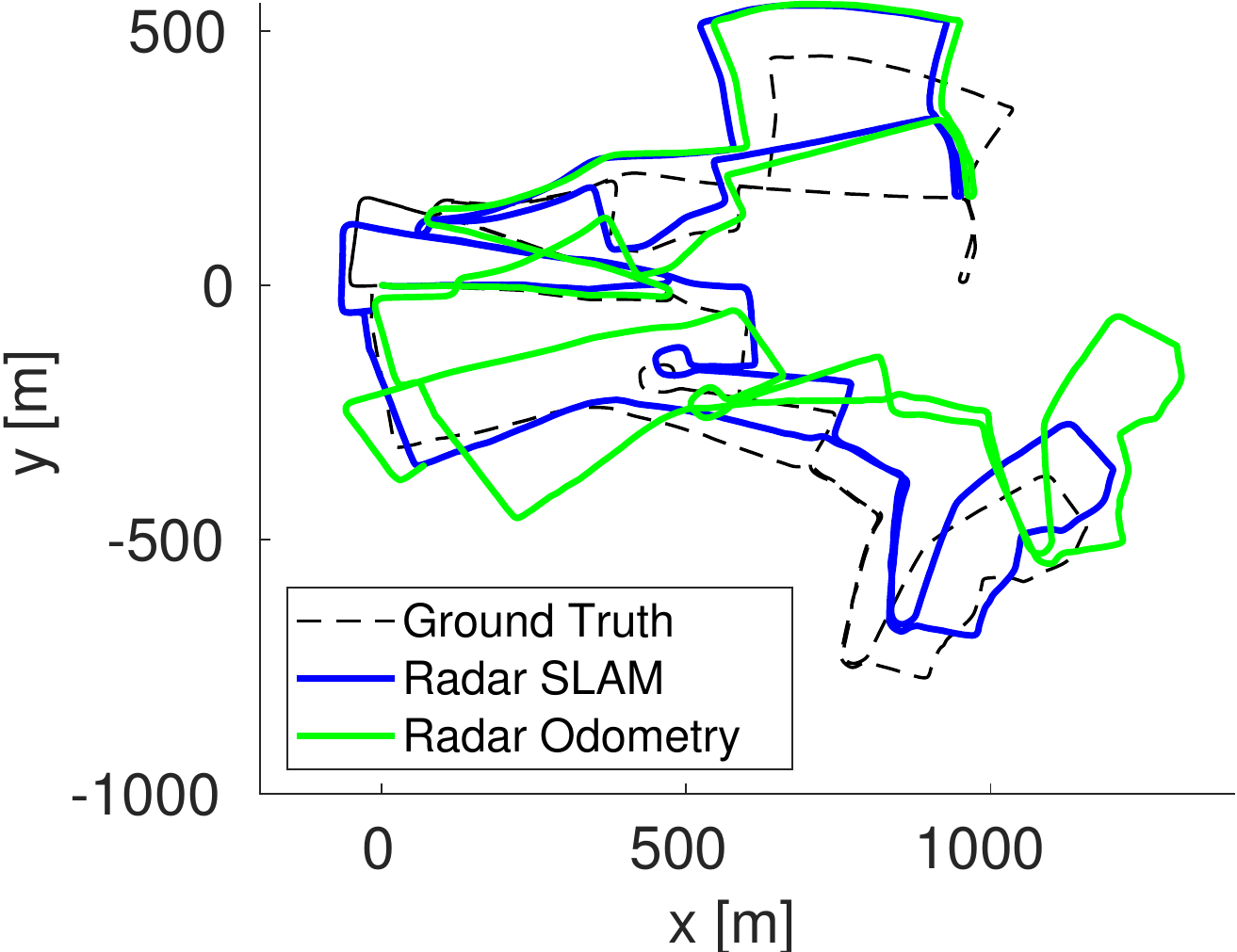}}\,
    % \hfill
    \subfloat[18-14-14-42]{\includegraphics[width=0.19\textwidth]{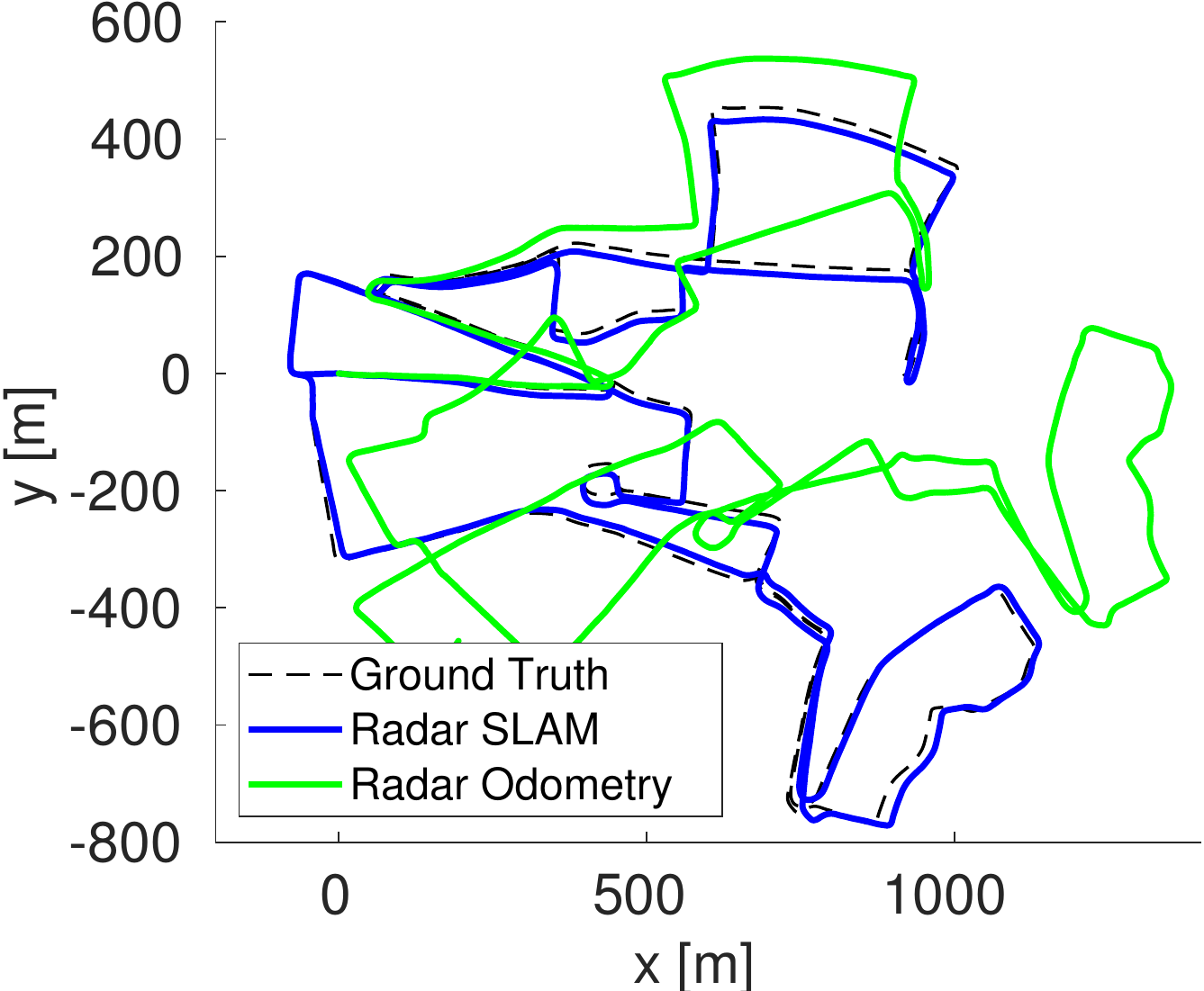}}\,
    % \hfill
    \subfloat[18-15-20-12]{\includegraphics[width=0.19\textwidth]{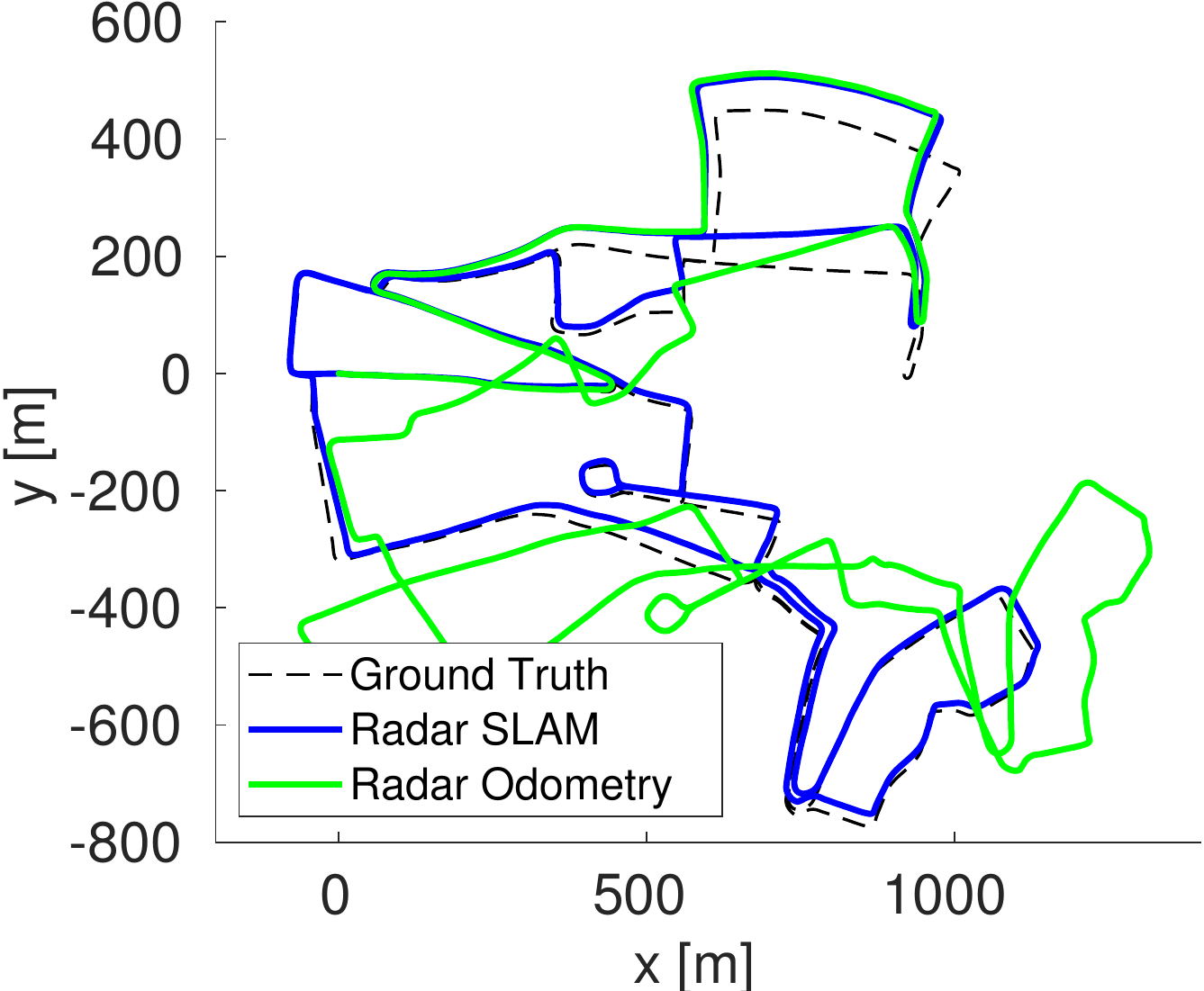}}
    \caption{Estimated Trajectories and Ground Truth of 5 Sequences from Oxford Radar RobotCar Dataset.}
  \label{fig:ours_method}
\end{figure*}

% \begin{figure*}
%     \begin{center}
%     \begin{subfigure}{0.23\linewidth}
%     \includegraphics[width=4.1cm,height=2.5cm]{figures/Result/google_foggy6.jpg}
%     \caption{Fog/Rain 1}
%   \end{subfigure}
%   \begin{subfigure}{0.185\linewidth}
%     \includegraphics[width=3.3cm,height=2.5cm]{figures/Result/google_foggy9.jpg}
%     \caption{Fog/Rain 2}
%   \end{subfigure}
%   \begin{subfigure}{0.185\linewidth}
%     \includegraphics[width=3.3cm,height=2.5cm]{figures/Result/google_snow_loop.jpg}
%     \caption{Snow}
%   \end{subfigure}
%   \begin{subfigure}{0.185\linewidth}
%     \includegraphics[width=3.3cm,height=2.5cm]{figures/Result/google_university_loop.jpg}
%     \caption{Countryside}
%   \end{subfigure}
%   \begin{subfigure}{0.185\linewidth}
%     \includegraphics[width=3.3cm,height=2.5cm]{figures/Result/google_night_loop.jpg}
%     \caption{Night}
%   \end{subfigure}
% \end{center}
%   \caption{Estimated Trajectories of Extreme Weather Sequences on Google Map.}
% \label{fig:google_plot}
% \end{figure*}
\begin{figure*}
\centering
    \subfloat[Fog/Rain 1]
    {\includegraphics[width=0.21\textwidth,height=2.5cm]{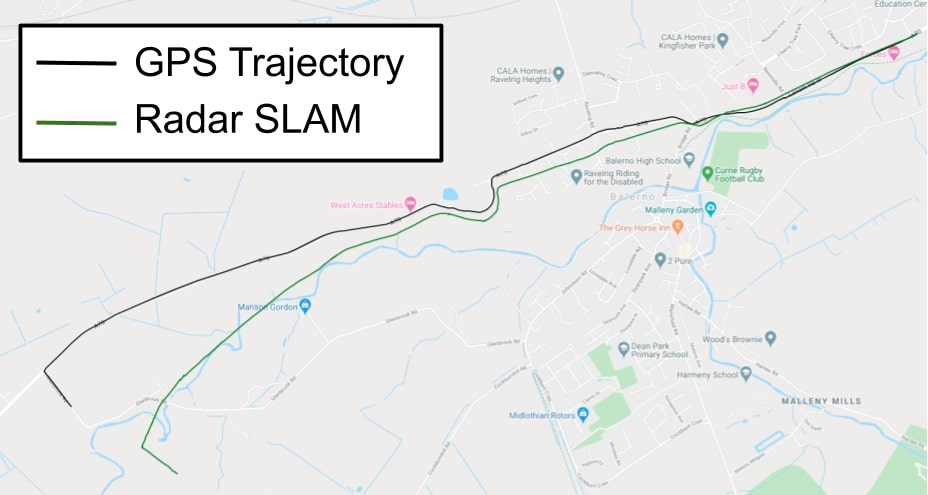}}
    \hfill
    \subfloat[Fog/Rain 2]
    {\includegraphics[width=0.185\textwidth,height=2.5cm]{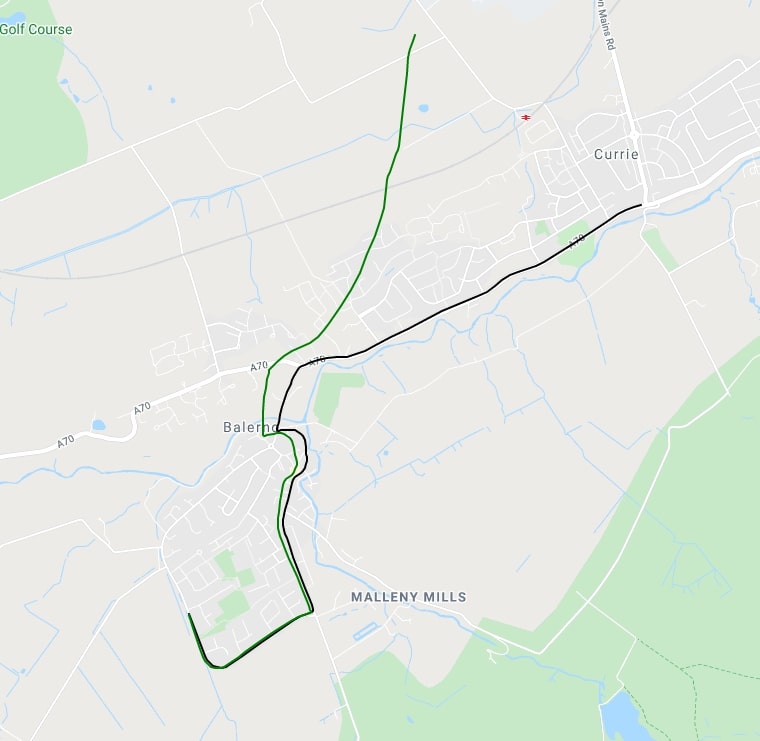}}
    \hfill
    \subfloat[Snow]
    {\includegraphics[width=0.185\textwidth,height=2.5cm]{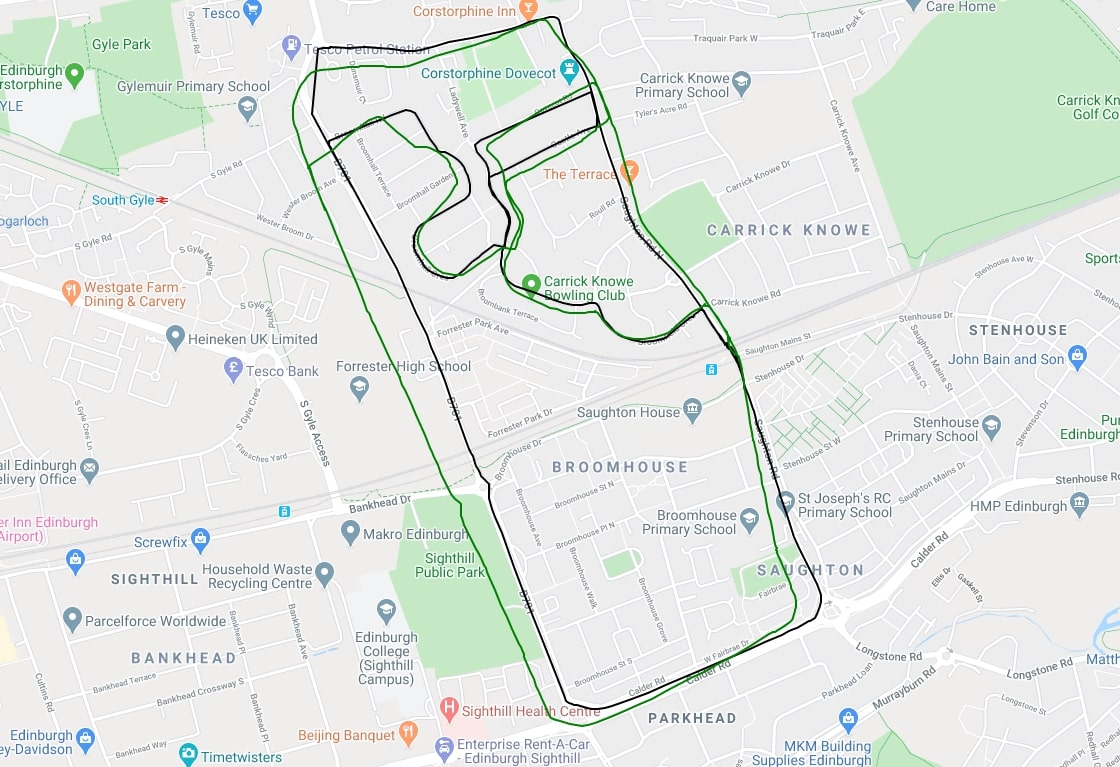}}
    \hfill
    \subfloat[Countryside]
    {\includegraphics[width=0.185\textwidth,height=2.5cm]{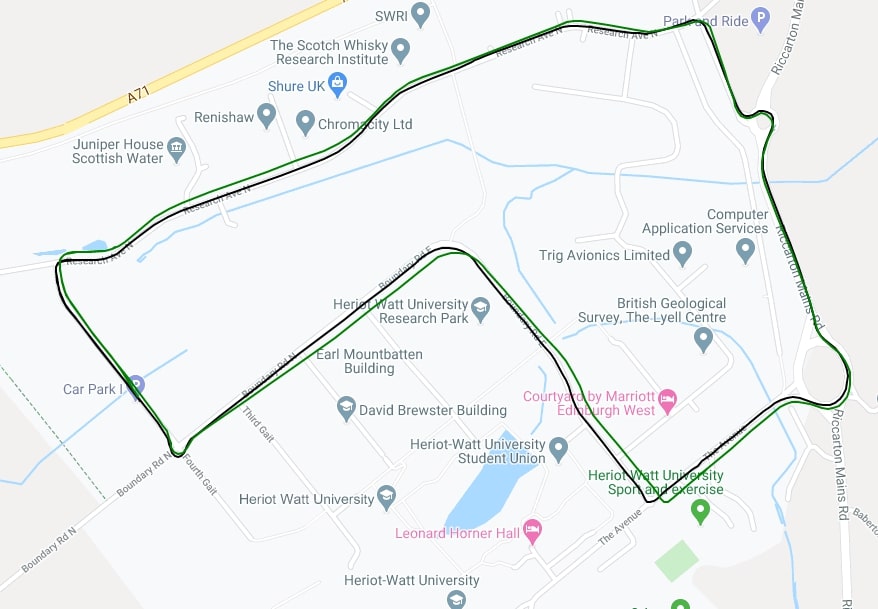}}
    \hfill
    \subfloat[Night]
    {\includegraphics[width=0.185\textwidth,height=2.5cm]{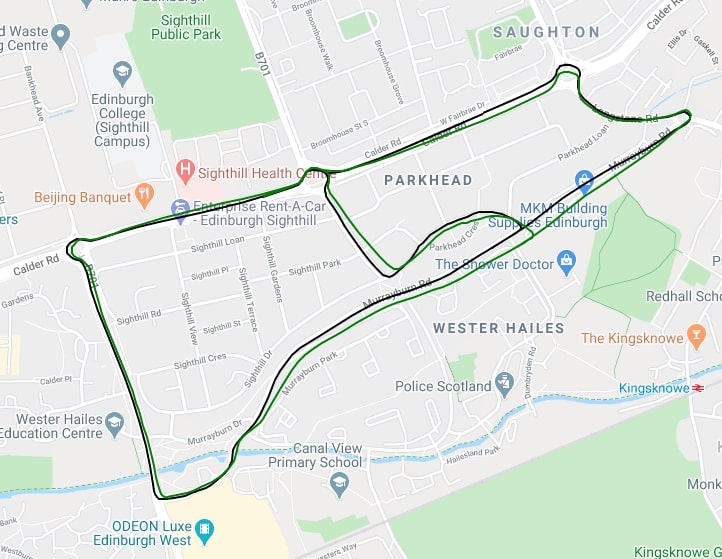}}

    \caption{Estimated Trajectories of Extreme Weather Sequences on Google Map.}
\label{fig:google_plot}
\end{figure*}

\subsection{Pose Graph Optimization}
A pose graph is gradually built as the radar moves. Once a loop is detected, the relative transformation between the current frame and the detected keyframe is computed by using ICP \cite{icp} with RANSAC \cite{ransac} as a geometric test, and added in the pose graph as a loop closure constraint. If the ICP converges, we then perform pose graph optimization on all the keyframes. We use g2o \cite{g2o} library for the pose graph optimization.
% We perform up to 10 Gauss-Newton iterations for the graph optimization, early stop if the epsilon is small enough.
After successfully optimizing the poses of the keyframes, we update all the map points for a global map.

\section{Experimental Results}\label{sec:exp_results}

Both quantitative and qualitative experiments are conducted to evaluate the proposed RadarSLAM in large-scale environments and some adverse weather conditions.

\subsection{Quantitative Evaluation}

The quantitative evaluation is to understand the pose estimation accuracy of the proposed RadarSLAM system. We follow the popular KITTI odometry evaluation criteria, i.e., computing the mean translation and rotation errors from length 100 to 800 meters with a 100m increment. State-of-the-art odometry and SLAM algorithms using different sensors are also compared.

The Oxford Radar RobotCar Dataset \cite{RadarRobotCarDatasetICRA2020} is used for the quantitative evaluation as it is an open large-scale radar dataset easy for benchmarking. 
% and has challenging dynamics, e.g., cars, buses and pedestrians, in an urban environment. 
It includes 32 sequences of radar data collected while traversing a same route in Oxford with ground truth poses. The radar data is captured by a Navtech CTS350-X, a Frequency Modulated Continuous Wave (FMCW) scanning radar. It is configured to return 3768 power readings at a resolution of 4.32cm across 400 azimuths and operates at the frequency of 4Hz (with maximum range of 163m). Due to the page limitation, the localization results of only 5 sequences are given in Table \ref{tab:result_oxford} and Fig. \ref{fig:ours_method}.

\noindent\textbf{Comparison with State-of-the-art Radar Odometry.}
Two state-of-the-art radar based pose estimation algorithms \cite{cen2018precise,barnes2019masking} using 360\degree FMCW radar are chosen for comparison. One is based on radar geometry and point clouds extracted from the radar images \cite{cen2018precise}, while the other one is a deep learning based approach \cite{barnes2019masking}. Since neither of them has an open-source implementation and the results reported in their papers only include mean errors, there is no detailed errors for each sequence in Table \ref{tab:result_oxford}. Their mean errors reported in \cite{barnes2019masking} are directly referenced here. It can be seen that the proposed radar odoemtry and SLAM methods both outperform the approach in \cite{cen2018precise} although the they are inferior to \cite{barnes2019masking}. Our radar SLAM system achieves 2.1854\% error in terms of translation and 0.0071 deg/m for rotation, which is much more accurate than the results of our pure odometry method.

% The first work for radar odometry  in the context of using 360\degree FMCW radar scanner proposed to extract landmark at each azimuth scan and associate two point cloud by finding the largest subsets of two pointclouds that share a similar shape. The state-of-the-art radar odometry method proposed in \cite{barnes2019masking} achieves excellent result by training a deep network to mask out noise and distractor objects.

% There is In terms of feature-based radar odometry, our method outperform the state-of-the-art method proposed by Cen $et$ $al$. \cite{cen2018precise}.

\noindent\textbf{Comparison with Camera and LiDAR Based Methods.}
Vision and LiDAR based algorithms are compared for comprehensive evaluation of the robustness and accuracy of RadarSLAM. ORB-SLAM 2 stereo \cite{murTRO2015}, LOAM \cite{loamZhang} and SuMa \cite{behley2018rss} fail to finish the whole sequences or drift fast due to the severe dynamics. Hence, their results in Table \ref{tab:result_oxford} are reported up to the point where they lose tracking. Note that the vision and LiDAR approaches provide 6 degrees of freedom poses which are projected onto the XY plane for evaluation. It can be seen that the proposed RadarSLAM can achieve comparable or better localization accuracy with enhanced robustness.

\begin{figure}
    % \begin{subfigure}[b]{0.32\columnwidth}
    % \includegraphics[width=\columnwidth]{figures/SnowRadar/000984.jpg}
    % \end{subfigure}
    % \begin{subfigure}[b]{0.32\columnwidth}
    % \includegraphics[width=\columnwidth]{figures/SnowRadar/000483.jpg}
    % \end{subfigure}
    % \begin{subfigure}[b]{0.32\columnwidth}
    % \includegraphics[width=\columnwidth]{figures/SnowRadar/005213.jpg}
    % \end{subfigure}
    % \vspace{0.1cm}

    % \begin{subfigure}[b]{0.49\columnwidth}
    % \includegraphics[width=\columnwidth,height=3.0cm]{figures/SnowRadar/road_snow.jpg}
    % \end{subfigure}
    % \begin{subfigure}[b]{0.50\columnwidth}
    % \includegraphics[width=\columnwidth,height=3.0cm]{figures/SnowRadar/sensor_snow.jpg}
    % \end{subfigure}
    \includegraphics[width=\columnwidth]{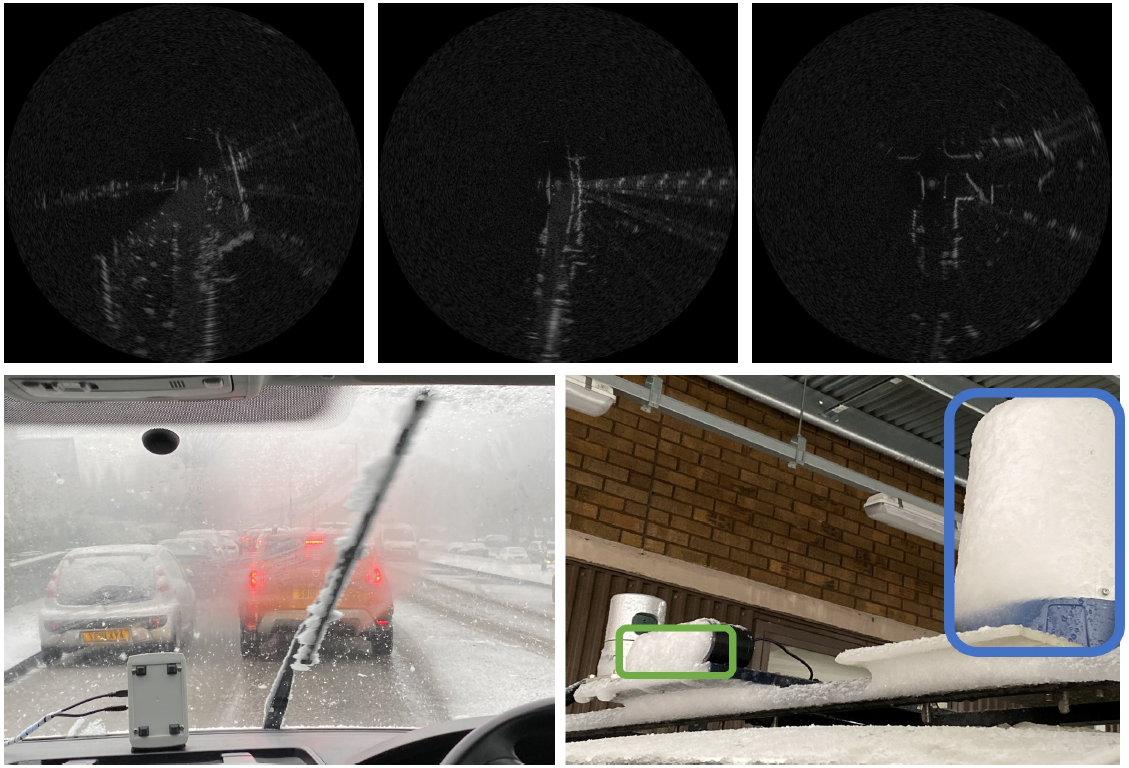}    
    
    \caption{Snow Sequence. Top: Radar images captured in snow. Note the front half of the radar data is lost due to thick snow covered on the radar. Bottom left: Photo of the heavy snowfall during data collection. Bottom right: Significant amount of snow covered on the camera, LiDAR and radar. }
    \label{fig:self_image}
\end{figure}

\begin{figure}
    % \begin{subfigure}[b]{0.32\columnwidth}
    % \includegraphics[width=\columnwidth,height=1.3cm]{figures/Fog/004178.jpg}
    % \end{subfigure}
    % \begin{subfigure}[b]{0.32\columnwidth}
    % \includegraphics[width=\columnwidth,height=1.3cm]{figures/Fog/000867.jpg}
    % \end{subfigure}
    % \begin{subfigure}[b]{0.32\columnwidth}
    % \includegraphics[width=0.32\columnwidth,height=1.3cm]{figures/Fog/000037.jpg}
    % \end{subfigure}
    % \vspace{0.1cm}

    % \begin{subfigure}[b]{0.32\columnwidth}
    % \includegraphics[width=0.32\columnwidth,height=1.3cm]{figures/Night/003039.jpg}
    % \end{subfigure}
    % \begin{subfigure}[b]{0.32\columnwidth}
    % \includegraphics[width=0.32\columnwidth,height=1.3cm]{figures/Night/000285.jpg}
    % \end{subfigure}
    % \begin{subfigure}[b]{0.32\columnwidth}
    % \includegraphics[width=0.32\columnwidth,height=1.3cm]{figures/Night/000054.jpg}
    % \end{subfigure}
    % \vspace{0.1cm}

    % \begin{subfigure}[b]{0.32\columnwidth}
    % \includegraphics[width=0.32\columnwidth,height=1.3cm]{figures/Snow/000014.jpg}
    % \end{subfigure}
    % \begin{subfigure}[b]{0.32\columnwidth}
    % \includegraphics[width=0.32\columnwidth,height=1.3cm]{figures/Snow/001316.jpg}
    % \end{subfigure}
    % \begin{subfigure}[b]{0.32\columnwidth}
    % \includegraphics[width=0.32\columnwidth,height=1.3cm]{figures/Snow/002796.jpg}
    % \end{subfigure}

    \includegraphics[width=\columnwidth,height=3.9cm]{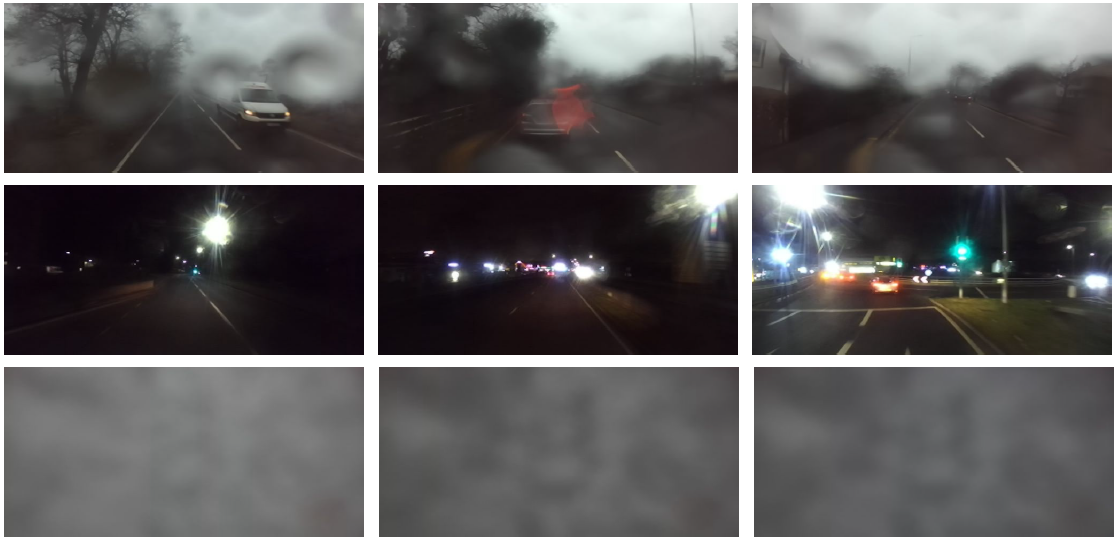}
    \caption{Images Collected in Fog/Rain (top), Night (middle) and Snow (bottom). Images quality significantly degrades in these conditions, making it extremely challenging for vision based algorithms. Note for the snow sequence at bottom, the camera is completely occluded by the heavy snowfall.}    
    \label{fig:foggy_image}
\end{figure}

\subsection{Qualitative Evaluation}
% To verify the robustness of the proposed radar system against other sensor modalities in adverse weathers, 5 radar sequences are self-collected in challenging weather conditions, including fog/rain, snow and night \footnote{These datasets will be open-source for the community.}. Our data collection vehicle is equipped with a GPS/IMU navigation system, a Velodyne HDL-32, a ZED stereo camera and a NavTech radar scanner. Our NavTech radar, whose maximum range is 100m and distance resolution is 0.1736 pixel/m, is slightly different from the one in Oxford Radar Dataset. The sequence lengths are given in Table \ref{tab:dataset}. Note that the Snow, Countryside and Night sequences have loop closure, while the other two sequences in fog do not have loops. Some sample camera images of fog/rain, night and snow sequences are shown in Fig. \ref{fig:foggy_image}, which show the significant challenges for vision based methods. The Snow sequence is particularly challenging with thick snow covering big portions of the camera, LiDAR and radar sensors as shown in Fig. \ref{fig:self_image}.

To verify the robustness of the proposed radar system against other sensor modalities in adverse weathers, 5 radar sequences are self-collected in challenging weather conditions, including fog/rain, snow and night \footnote{For more details and dataset, please visit: 
\textcolor{blue}{\url{http://pro.hw.ac.uk/radarslam}
}}. Our data collection vehicle is equipped with a GPS/IMU navigation system, a Velodyne HDL-32, a ZED stereo camera and a NavTech radar scanner. Our NavTech radar, whose maximum range is 100m and distance resolution is 0.1736 pixel/m, is slightly different from the one in Oxford Radar Dataset. The sequence lengths are given in Table \ref{tab:dataset}. Note that the Snow, Countryside and Night sequences have loop closure, while the other two sequences in fog do not have loops. Some sample camera images of fog/rain, night and snow sequences are shown in Fig. \ref{fig:foggy_image}, which show the significant challenges for vision based methods. The Snow sequence is particularly challenging with thick snow covering big portions of the camera, LiDAR and radar sensors as shown in Fig. \ref{fig:self_image}.

\begin{table}
\footnotesize
\caption{Sequence Length of Self-collected Dataset}\label{tab:dataset}
\scalebox{0.9}{
\begin{tabular}{c|c|c|c|c|c}
\hline
Sequence    & Fog/Rain 1 & Fog/Rain 2 & Snow & Countryside & Night \\ \hline
Length (km) & 4.7   & 4.8   & 8.7  & 3.4        & 5.6   \\ \hline
\end{tabular}
}
\end{table}

The estimated trajectories of the sequences collected in the extreme weathers are shown in Fig. \ref{fig:google_plot} plotted on Google map. The pose estimates of the two Fog/Rain sequences drift over time as there is no loop, while the results of the Snow, Countryside and Night are close to the ground truth. In contrast, the LiDAR based methods are affected by the fog and snow. Since the camera is blocked by the water drops in the foggy/rainy weather, vision based approaches also fail. In the night sequence, the camera images have serious motion blur, see Fig. \ref{fig:foggy_image}. The Snow sequence is the most challenging one since the heavy snowfall causes occlusion to all the sensors, especially the camera. Among all the three sensor modalities, only the radar system is able to operate and localize reliably in all the weather conditions.

% \begin{figure}[h]
%     \centering
%     \includegraphics[width=\columnwidth]{figures/Result/map_snow.jpg}
%     \caption{Mapping result on \textbf{Snow} sequence.
%     }
%     \label{fig:map_snow}
% \end{figure}

% \subsection{Ablation Study of Proposed System}
% \subsubsection{Cartesian Image Resolution}
% \subsubsection{Understand the Drift Source}

\begin{figure}
    \centering
    \includegraphics[width=\columnwidth]{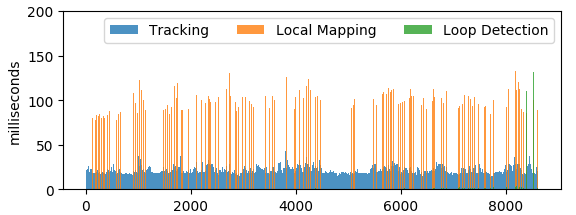}
    \caption{Runtime on Sequence 10-12-32-52 of Oxford Radar RobotCar Dataset.}
    \label{fig:runtime}
\end{figure}
\subsection{Computation Time}
The RadarSLAM system is implemented in C++ without GPU. The whole system runs at $\sim6$Hz on a laptop with an Intel i7 2.60GHz CPU and 16 GB RAM. The runtime of a 37-minute sequence (about 9000 images) is presented in Fig. \ref{fig:runtime}, demonstrating its real-time performance.

\section{Conclusion}
In this work, a full SLAM system is designed for FMCW radar that is able to reliably operate in large scale environment online and construct globally consistent maps. An effective probabilistic model of point cloud generation is proposed for loop closure detection. Extensive experiments are conducted on the publicly available Oxford radar dataset and the self-collected datasets in adverse weathers. The future work will investigate the fusion of Inertial Measurement Unit with radar for better accuracy.
% Second, how to perform preprocessing to remove noises like speckle noise and multipath reflection by estimating the signal statistics. 
% Third, how to leverage the structural information in the radar scan and construct a descriptor for loop closure detection even when the robot reversely visits the same place.

\section*{Acknowledgements}
We thank Saptarshi Mukherjee, Marcel Sheeny and Dr. Yun Wu for the help of our data collection. 
We thank Raul Mur-Artal for making ORB-SLAM code open-source, which greatly inspires our implementation.

% \bibliographystyle{IEEEtran}
% \bibliography{references}

\begin{thebibliography}{10}
\providecommand{\url}[1]{#1}
\csname url@samestyle\endcsname
\providecommand{\newblock}{\relax}
\providecommand{\bibinfo}[2]{#2}
\providecommand{\BIBentrySTDinterwordspacing}{\spaceskip=0pt\relax}
\providecommand{\BIBentryALTinterwordstretchfactor}{4}
\providecommand{\BIBentryALTinterwordspacing}{\spaceskip=\fontdimen2\font plus
\BIBentryALTinterwordstretchfactor\fontdimen3\font minus
  \fontdimen4\font\relax}
\providecommand{\BIBforeignlanguage}[2]{{%
\expandafter\ifx\csname l@#1\endcsname\relax
\typeout{** WARNING: IEEEtran.bst: No hyphenation pattern has been}%
\typeout{** loaded for the language `#1'. Using the pattern for}%
\typeout{** the default language instead.}%
\else
\language=\csname l@#1\endcsname
\fi
#2}}
\providecommand{\BIBdecl}{\relax}
\BIBdecl

\bibitem{marck2013indoor}
J.~W. Marck, A.~Mohamoud, E.~vd~Houwen, and R.~van Heijster, ``Indoor radar
  slam a radar application for vision and gps denied environments,'' in
  \emph{2013 European Radar Conference}.\hskip 1em plus 0.5em minus 0.4em\relax
  IEEE, 2013.

\bibitem{vivet2013localization}
D.~Vivet, P.~Checchin, and R.~Chapuis, ``Localization and mapping using only a
  rotating fmcw radar sensor,'' \emph{Sensors}, 2013.

\bibitem{schuster2016landmarkRadar}
F.~Schuster, C.~G. Keller, M.~Rapp, M.~Haueis, and C.~Curio, ``Landmark based
  radar {SLAM} using graph optimization,'' in \emph{IEEE International
  Conference on Intelligent Transportation Systems}.\hskip 1em plus 0.5em minus
  0.4em\relax IEEE, 2016.

\bibitem{holder2019realRadarGraph}
M.~Holder, S.~Hellwig, and H.~Winner, ``Real-time pose graph slam based on
  radar,'' in \emph{IEEE Intelligent Vehicles Symposium}.\hskip 1em plus 0.5em
  minus 0.4em\relax IEEE, 2019.

\bibitem{cen2018precise}
S.~Cen and P.~Newman, ``Precise ego-motion estimation with millimeter-wave
  radar under diverse and challenging conditions,'' in \emph{IEEE International
  Conference on Robotics and Automation}.\hskip 1em plus 0.5em minus
  0.4em\relax IEEE, 2018, pp. 1--8.

\bibitem{cen2019radar}
S.~H. Cen and P.~Newman, ``Radar-only ego-motion estimation in difficult
  settings via graph matching,'' in \emph{International Conference on Robotics
  and Automation}.\hskip 1em plus 0.5em minus 0.4em\relax IEEE, 2019, pp.
  298--304.

\bibitem{aldera2019Introspective}
R.~Aldera, D.~De~Martini, M.~Gadd, and P.~Newman, ``What could go wrong?
  introspective radar odometry in challenging environments,'' in \emph{IEEE
  Intelligent Transportation Systems Conference}.\hskip 1em plus 0.5em minus
  0.4em\relax IEEE, 2019.

\bibitem{PhaRaO}
Y.~S. Park, Y.-S. Shin, and A.~Kim, ``Pharao: Direct radar odometry using phase
  correlation,'' in \emph{Proceedings of the IEEE International Conference on
  Robotics and Automation (ICRA)}.\hskip 1em plus 0.5em minus 0.4em\relax IEEE,
  May 2020.

\bibitem{RadarRobotCarDatasetICRA2020}
D.~Barnes, M.~Gadd, P.~Murcutt, P.~Newman, and I.~Posner, ``The oxford radar
  robotcar dataset: A radar extension to the oxford robotcar dataset,'' in
  \emph{IEEE International Conference on Robotics and Automation}, 2020.

\bibitem{barnes2019masking}
D.~Barnes, R.~Weston, and I.~Posner, ``Masking by moving: Learning
  distraction-free radar odometry from pose information,'' \emph{arXiv preprint
  arXiv:1909.03752}, 2019.

\bibitem{aldera2019fast}
R.~Aldera, D.~De~Martini, M.~Gadd, and P.~Newman, ``Fast radar motion
  estimation with a learnt focus of attention using weak supervision,'' in
  \emph{2019 International Conference on Robotics and Automation (ICRA)}.\hskip
  1em plus 0.5em minus 0.4em\relax IEEE, 2019, pp. 1190--1196.

\bibitem{UnderTheRadarICRA2020}
D.~Barnes and I.~Posner, ``Under the radar: Learning to predict robust
  keypoints for odometry estimation and metric localisation in radar,'' in
  \emph{IEEE International Conference on Robotics and Automation}, 2020.

\bibitem{millimeterFMCW}
M.~Skolnik, ``Radar handbook,'' 1970.

\bibitem{cen2019ego}
S.~Cen, ``Ego-motion estimation and localization with millimeter-wave scanning
  radar,'' Master's thesis, University of Oxford, 2019.

\bibitem{bay2008speeded}
H.~Bay, A.~Ess, T.~Tuytelaars, and L.~Van~Gool, ``Speeded-up robust features
  (surf),'' \emph{Computer vision and image understanding}, vol. 110, no.~3,
  pp. 346--359, 2008.

\bibitem{challis1995procedure}
J.~H. Challis, ``A procedure for determining rigid body transformation
  parameters,'' \emph{Journal of Biomechanics}, vol.~28, pp. 733--737, 1995.

\bibitem{murTRO2015}
M.~J. M.~M. Mur-Artal, Ra\'ul and J.~D. Tard\'os, ``{ORB-SLAM}: a versatile and
  accurate monocular {SLAM} system,'' \emph{IEEE Transactions on Robotics},
  vol.~31, no.~5, pp. 1147--1163, 2015.

\bibitem{bundle}
B.~Triggs, P.~F. McLauchlan, R.~I. Hartley, and A.~W. Fitzgibbon, ``Bundle
  adjustment—a modern synthesis,'' in \emph{International Workshop on Vision
  Algorithms}.\hskip 1em plus 0.5em minus 0.4em\relax Springer, 1999, pp.
  298--372.

\bibitem{churchill2012experience}
W.~Churchill, ``Experience based navigation: Theory, practice and
  implementation,'' 2012.

\bibitem{loamZhang}
J.~Zhang and S.~Singh, ``Loam: Lidar odometry and mapping in real-time,'' in
  \emph{Proceedings of Robotics: Science and Systems Conference}, 2014.

\bibitem{behley2018rss}
J.~Behley and C.~Stachniss, ``Efficient surfel-based {SLAM} using {3D} laser
  range data in urban environments,'' in \emph{Robotics: Science and Systems},
  2018.

\bibitem{he2016m2dp}
L.~He, X.~Wang, and H.~Zhang, ``M2dp: A novel 3d point cloud descriptor and its
  application in loop closure detection,'' in \emph{IEEE/RSJ International
  Conference on Intelligent Robots and Systems}.\hskip 1em plus 0.5em minus
  0.4em\relax IEEE, 2016, pp. 231--237.

\bibitem{icp}
P.~J. Besl and N.~D. McKay, ``Method for registration of 3-d shapes,'' in
  \emph{Sensor fusion IV: control paradigms and data structures}, vol.
  1611.\hskip 1em plus 0.5em minus 0.4em\relax International Society for Optics
  and Photonics, 1992, pp. 586--606.

\bibitem{ransac}
M.~A. Fischler and R.~C. Bolles, ``Random sample consensus: a paradigm for
  model fitting with applications to image analysis and automated
  cartography,'' \emph{Communications of the ACM}, 1981.

\bibitem{g2o}
R.~K{\"u}mmerle, G.~Grisetti, H.~Strasdat, K.~Konolige, and W.~Burgard, ``g2o:
  A general framework for graph optimization,'' in \emph{IEEE International
  Conference on Robotics and Automation}.\hskip 1em plus 0.5em minus
  0.4em\relax IEEE, 2011, pp. 3607--3613.

\end{thebibliography}

\end{document}